# Enhancing Building Safety Design for Active Shooter Incidents: Exploration of Building Exit Parameters using Reinforcement Learning-Based Simulations

Ruying Liu, Wanjing Wu, Burcin Becerik-Gerber, Gale M. Lucas
University of Southern California, United States
ruyingli@usc.edu

**Abstract.** With the alarming rise in active shooter incidents (ASIs) in the United States, enhancing public safety through building design has become a pressing need. This study proposes a reinforcement learning-based simulation approach addressing gaps in existing research that has neglected the dynamic behaviours of shooters. We developed an autonomous agent to simulate an active shooter within a realistic office environment, aiming to offer insights into the interactions between building design parameters and ASI outcomes. A case study is conducted to quantitatively investigate the impact of building exit numbers (total count of accessible exits) and configuration (arrangement of which exits are available or not) on evacuation and harm rates. Findings demonstrate that greater exit availability significantly improves evacuation outcomes and reduces harm. Exits nearer to the shooter's initial position hold greater importance for accessibility than those farther away. By encompassing dynamic shooter behaviours, this study offers preliminary insights into effective building safety design against evolving threats.

## 1. Introduction

The occurrence of active shooter incidents (ASIs) in the United States has alarmingly increased in recent years, posing a substantial risk to public safety and human lives. Consequently, designing buildings that offer a safer environment for occupants during emergencies is crucial for reducing harm and facilitating safer evacuation. Empirical studies and computer simulations have played a key role in identifying critical design factors for civilian evacuation, including the positions and sizes of doors, exits, and obstacles (Gao et al., 2020).

However, significant gaps remain in current understanding of building safety design, especially in the context of ASIs. There is a lack of research on how building design parameters affect ASI outcomes. Few studies have investigated this challenge, and their findings primarily rely on civilian simulations, without considering the perspective of the shooter's behavior (Arteaga and Park, 2020; Zhu et al., 2023). For example, design elements like curved hallways and wing walls, hypothesized to limit a shooter's line of sight and reduce casualties, have not been adequately tested. These gaps underscore the critical need for a more realistic simulation of shooter's behaviors. While there have been studies in the domain of ASI simulations, a major limitation has been their reliance on predefined shooter paths. Such an approach provides limited insight and fails to capture the intricate behaviors of shooters under realistic and complex scenarios (Arteaga and Park, 2020).

To address these challenges, the present study introduces a novel approach based on Reinforcement Learning (RL) to construct intelligent agents capable of simulating active shooters within complex built environments. The motivation behind this endeavor involves two aspects: firstly, to assess safety design parameters in relation to shooter behavior in ASIs, thus, to discover hidden vulnerabilities within built environments; and secondly, to overcome the limitations of existing simulation techniques by expanding the scope beyond pre-defined paths to include more realistic behaviors and dynamic contexts. By utilizing reinforcement learning, our method enables the creation of agents that autonomously perceive their environment and make informed decisions based on a wide range of possible scenarios. The remainder of this paper is organized as follows. Section 2 explains the methodology on training the autonomous shooter agent, including the workflow of the RL model, configuration settings, reward functions, the simulated office environment, and employed training strategies. Section 3 presents the case study, where simulations quantitatively assess the impacts of the number of exits (the total count of accessible exits) and configurations of exits (the arrangement of which exits are available or not) on building safety measured by evacuation rate and harm rate. Section 4 reports the simulation results and discusses the findings. Section 5 summarizes the findings.

## 2. Methodology

### 2.1 Reinforcement Learning Network

Different from studying building occupants' behavior through human-subject experiments, it is challenging to collect real-world data on shooter behavior due to ethical concerns. While understanding the characteristics of active shooters is complex, certain objectives of active shooters have been identified through prior research and law enforcement reports. A shooter usually has an objective to punish the public by maximizing the number of casualties (Fox and DeLateur, 2014), which can be represented through RL. The goal of the agent is to select actions that maximize the cumulative future reward, formulated as:

$$V_{\pi(s)} = E[R_{t+1} + \gamma R_{t+2} + \gamma^2 R_{t+3} + \cdots | S_t = s, \pi] \quad (1)$$

where $V_{\pi(s)}$ is the expected cumulative reward when following policy $\pi$ from state s, $R_{t+n}$ is the reward at time step t+n, and $\gamma$ is the discount factor with a value between 0 and 1.

To model shooter's behavior, we have implemented a deep RL network based on the Proximal Policy Optimization (PPO) algorithm, which facilitates the agent's learning for both navigation in complex environments and the achievement of its objective. Figure 1 illustrates the workflow of our model, presenting agent's operational dynamics from perceiving environmental cues to executing actions and receiving feedback. The Ray Perception Sensor 3D was utilized to process raw environmental data into a set of features, including velocity, position, obstacles, and targets through ray cast-based observations. These features constituted the observations that inform the agent's decision-making process, executed by the brain using the PPO algorithm, a highly efficient and popular algorithm proposed by OpenAI (Schulman et al., 2017). Agent interacted with the environment based on actions determined by the brain, leading to changes within the environment that generate rewards. These rewards are fed back to the brain to fine-tune the policy, creating a closed-loop system that drives learning and adaptation.

We utilized the Unity ML-Agent toolkit (Juliani et al., 2020) as our reinforcement learning platform. Decision-making processes were conducted using TensorFlow, with actions controlled through a Python API that interacts with the Unity environment via the external communicator. Table 1 outlines the configuration of our ML-Agent environment. The parameters were chosen through hyperparameter sweeps over the typical ranges of values for different hyperparameter provided by Unity ML-Agents' documentation. The maximum number of steps was set to 5,000,000 to provide the agent sufficient opportunity to learn from the environment through extensive interaction. A summary frequency of 50,000 steps provided a balance between computational overhead and the granularity of performance tracking. The network settings were configured to normalize inputs, with two layers of 128 hidden units each, selected for their ability to capture complex patterns in the data without overfitting.

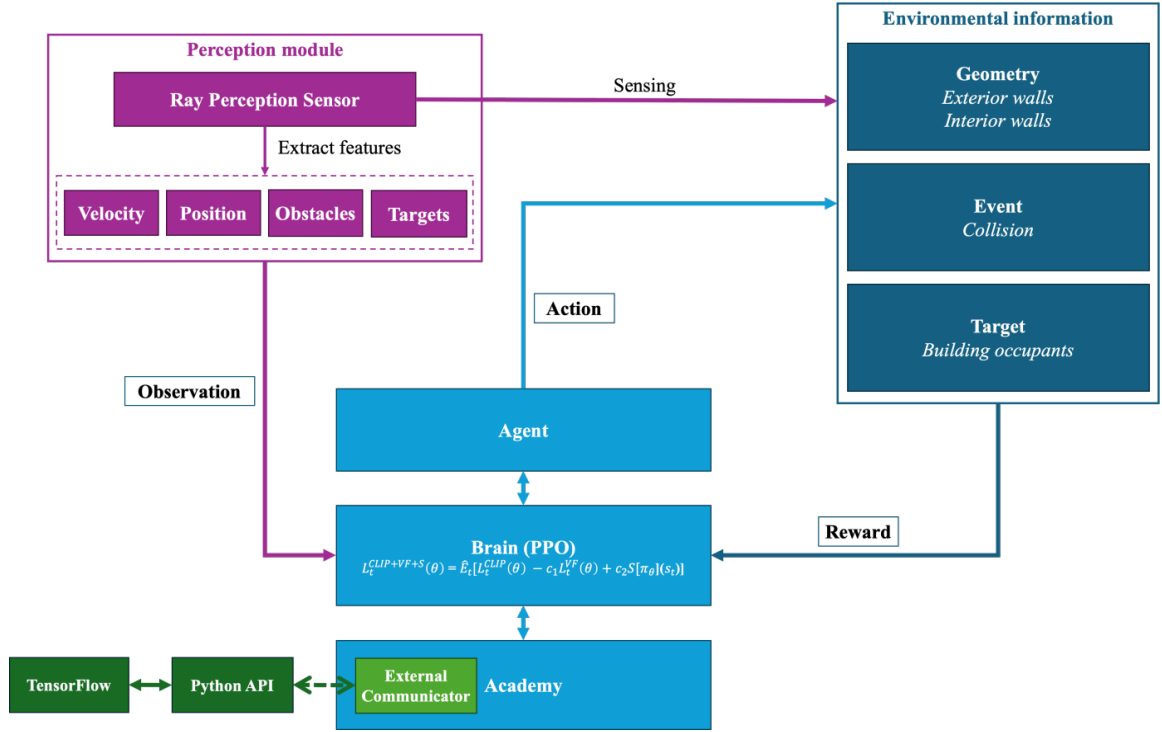

Figure 1: Integrated Workflow of a PPO-Based Reinforcement Learning Model

Table 1: Configuration settings of the ML-Agents environment

| Category | Environment | | | |
|---|---|---|---|---|
| Parameter | Max Steps | Time Horizon | Summary Frequency | Keep Checkpoints |
| Value | 5,000,000 | 64 | 50,000 | 100 |
| Category | Hyperparameter | | | |
| Parameter | Batch Size | Buffer Size | Learning Rate | Beta |
| Value | 2048 | 20480 | 0.0003 | 0.01 |
| Parameter | Epsilon | Lambda | Num Epoch | Learning Rate Schedule |
| Value | 0.2 | 0.95 | 3 | linear |
| Parameter | Beta Schedule | Epsilon Schedule | | |
| Value | linear | linear | | |
| Category | Network Setting | | | |
| Parameter | Normalize | Hidden Units | Number of Layers | Visualization Type |
| Value | true | 128 | 2 | simple |
| Parameter | Memory | Goal conditioning type | Deterministic | |
| Value | null | hyper | false | |
| Category | Reward Signal | | | |

| Parameter | Extrinsic Gamma | Extrinsic Strength | | |
|---|---|---|---|---|
| Value | 0.99 | 1.0 | | |

## 2.2 Reward Functions

Agent's behavior is dictated by reward signals through providing feedback on its actions. The reward system in this study is as

$$R = r_{exterior\ wall} + r_{interior\ wall} + r_{target} + r_{time} \tag{2}$$

R represents the total reward that the agent can receive. It is a linear combination of three award terms: $r_{target}$ refers to the reward if the shooter reaches a target (i.e., a building occupant); $r_{exterior\ wall}$ is a penalty if the shooter collides with an exterior wall; $r_{interior\ wall}$ refers to the penalty when the agent collides with an interior wall.

**Target reward**. $r_{target}$ motivates the agent to move towards building occupant targets. We set a linear function that correlates with number of targets the agent has reached so far. The agent receives a reward of 10 points increased by an additional 5 points for each target it has found so far. The more targets the agent has reached so far, the larger reward it will receive for reaching a new target. The escalating reward is because there will be less unfound targets in the environment as the agent progresses, causing an increased challenge of locating remaining targets over time. Therefore, this mechanism encourages continuation of the target-seeking behavior. Furthermore, it mimics the escalating urgency an actual shooter might experience to maximize target engagement before the imminent intervention of law enforcement authorities.

$$r_{target} = \begin{cases} 10 + count \times 5\ if\ reaches\ a\ target \\ 0\ otherwise \end{cases} \tag{3}$$

*where count is number of reached targets so far*

**Collision penalties**. To discourage the agent from colliding with walls, penalties are administered: a higher linear penalty for collisions with exterior walls, and a lesser fixed penalty for interior walls. Similar to the $r_{target}$, the $r_{exterior\ wall}$ is linearly increased with number of reached targets so far to encourage better navigation or less mistakes as time goes and as the shooter go deeper into the building. The shooter will receive the smallest penalty for colliding with the exterior wall at the beginning of the simulation to facilitate the movement. If the penalty is too large at the beginning, the shooter agent may not move or just circle at the starting location instead of searching for targets. For $r_{interior\ wall}$, given our complicated building layout, we implemented a relatively small penalty for it to avoid the penalty holding the agent from exploring the whole building.

$$r_{exterior\ wall} = \begin{cases} -2 - count \times 0.2\ if\ colliding\ with\ an\ exterior\ wall \\ 0\ otherwise \end{cases} \tag{4}$$

$$r_{interior\ wall} = \begin{cases} -0.5\ if\ colliding\ with\ an\ interior\ wall \\ 0\ otherwise \end{cases} \tag{5}$$

$$r_{time} = -0.001\ per\ step \tag{6}$$

**Time penalty**. Finally, a time penalty of -0.001 per step is applied to encourage the agent to complete its task efficiently, balancing exploration with the urgency of the mission.

### 2.3 Training Environment

The first floor of an office building model is utilized in this study (Figure 2). According to International Building Code (IBC) Table 1004.5 issued by International Code Council (ICC, 2021), the minimum floor area required per person for business occupancies (e.g., offices) is 150 ft$^2$ (i.e., 13.94 m$^2$). With our building's area being 2400 m$^2$, the maximum occupancy is calculated at approximately 172. To promote generalization of the trained model in varied and new environments, we employed randomization and reward shaping strategies. Specifically, we randomly altered agents' starting position to prevent overfitting to a specific path and to encourage exploration. In addition to varying the starting points, we respawned 60 occupants randomly over the building footprint at each training episode, satisfying the occupancy limit. For efficient training, we employed parallel computation by duplicating the training environment and setting multiple agents to a single brain to decide actions in a batch fashion. The ray perception sensor was configured with a maximum ray degree of 70 on each side of the center line, resulting in a total field of view of 140 degrees. In addition to a single ray pointing straightforward, three additional rays were emitted per side, leading to a total of seven rays. Each ray had a length of 20 meters, meaning that any objects beyond this distance would not be detected. When any of these rays intersected with a target within the 20-meter range, the sensor perceived the target and processed the information to determine the next movement. The shooting action was simulated using a collision detection mechanism with a threshold radius of 2.7 meters (equivalent to 3 yards). This indicates that any targets within the range would be considered harmed. This close range was selected to reflect the high shooting accuracy in confined spaces. The choice of 3 yards was also informed by the FBI pistol qualification course, which included shooting at distances starting from 3 yards (Morrison and Vila, 1998), reflecting a practical distance encountered in real shootings.

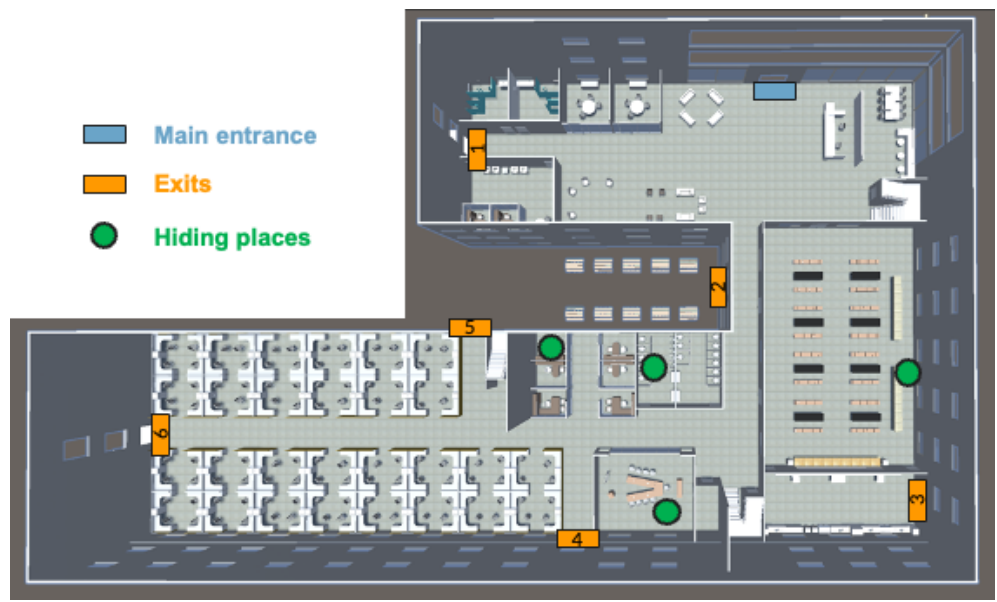

Figure 2: Building Layout

### 3. Case Study

Exits play a significant role in survival outcomes during building emergencies (Kobes et al., 2010). Upon the development of an autonomous shooter agent via RL, we conducted simulations to assess how the number and configuration of exits affect building safety during ASIs. The shooter's position, the total number of building occupants reaching the entrance and the exits, and the number harmed by the shooter were recorded in real time. We analyzed two variables: evacuation rate (the percentage of occupants who successfully exited the building before being reached by the shooter) and harm rate (the percentage of occupants reached by the shooter throughout a simulation run).

Our experiment utilized the same multi-room building as the training environment with a main entrance and 6 exits (Figure 2). To understand the influence of exit availability and location on evacuation effectiveness and casualty rates, we implemented a series of configurations ranging from all exits available to progressively fewer exits available. We conducted 100 simulation runs for each exit configuration, defined by the number and specific locations of exits. Within occupancy limit, each simulation initialized with 100 building occupants uniformly distributed at random throughout the building to reflect the unpredictability of real-life incidents. The shooter agent was randomly positioned within a predefined square area (4 units × 4 units) near the main entrance to reflect common entry points in real incidents and to introduce variability for robust findings across various conditions.

Controlled by the trained brain, the shooter agent (red cube in Figure 3) immediately began searching for visible targets as the simulation started. Building occupants, shown as green spheres in the same figure, delayed their response by 3 seconds to simulate the initial freeze response. Each occupant then navigated at a constant speed towards the closest hiding place, available exit, or main entrance (labeled in Figure 2), following the shortest distance strategy where the exit decision is driven by distance (Ma et al., 2021). The four hiding places (shown in Figure 2) were selected based on findings from previous human-subject experiments (Liu et al., 2023a; Liu et al., 2023b; Zhu et al., 2022), representing typical hiding locations. If reached by the shooter, an occupant would change color to red (shown in Figure 3), ceasing movement to indicate harm. Each episode ended if the shooter agent collided with any exterior wall or failed to locate a new target (i.e., a green sphere) within a 20-second window. This duration was determined from trial simulations, where agents not finding a target within 20 seconds typically engaged in aimless or repetitive actions, making further continuation inefficient.

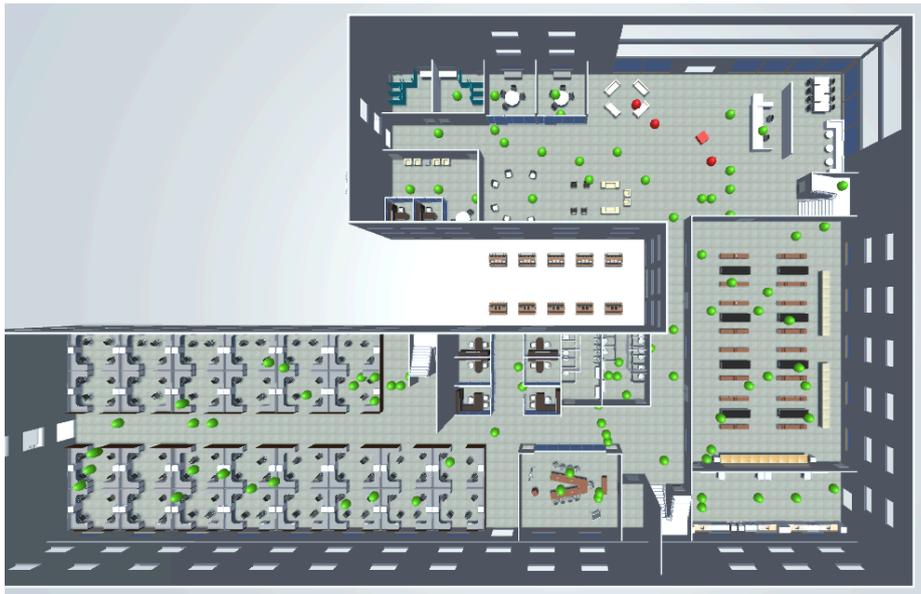

Figure 3: Early State of a Simulation (red cube: shooter agent, green sphere: unharmed building occupants, and red sphere: harmed building occupants)

In this case study, we examined three primary scenarios distinguished by the number of accessible exits. Full access is the baseline scenario with the entrance and all 6 exits available. Figure 4 shows the shooter agent's moving trajectories across 100 simulation runs in this scenario. The second scenario models the condition where one exit is unavailable, resulting in 6 subcases corresponding to each exit's closure. The third scenario further constrains the evacuation options by making two exits unavailable, for which we analyzed all 15 possible combinations of the remaining 4 available exits. All scenarios comply with the IBC (ICC, 2021), which requires at least two exits for buildings with fewer than 500 occupants. We conducted 100 simulation runs per subcase, resulting in sample sizes of 100 for the full access scenario, 600 for the scenario with one exit blocked, and 1500 for the scenario with two unavailable exits.

We used one-way ANOVAs to examine the effects of exit numbers (i.e., full access, five available exits, and 4 available exits) and exit configurations within each exit number category (i.e., 1 case for full access, 6 cases for five available exits, and 15 for 4 available exits). The sensitivity analysis done through G*Power 3.1 (Faul et al., 2007) with an alpha of 0.5 indicates that we have 80% power to detect small effects for ANOVA (f = 0.099).

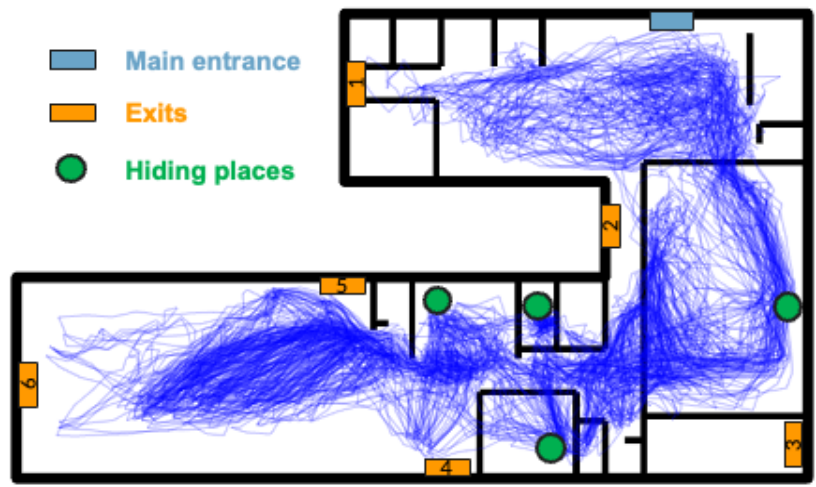

Figure 4: Moving Trajectories of the Shooter Agent in the Full Access Scenario (N = 100)

## 4. Results and Discussion

As shown in Figure 5, a significant effect of number of exits was found ($F(2, 2197) = 27.796$, $p < 0.001$, $\eta_p^2 = 0.025$, $d = 0.320$), indicating that full access scenario (M = 35.600%, SD = 10.852%) has the lowest harm rate compared to the scenarios with five available exits (M = 38.190%, SD = 12.501%), which in turn have a lower harm rate than those with four available exits (M = 42.249%, SD = 13.846%). Conversely, another significant effect of the number of exits was found ($F(2, 2197) = 177.466$, $p < 0.001$, $\eta_p^2 = 0.139$, $d = 0.804$), showing that the full access scenario (M = 50.850%, SD = 7.651%) had the highest evacuation rate, followed by five available exits (M = 46.058%, SD = 6.982%), and then by four available exits (M = 40.793%, SD = 7.385%). These findings suggest a correlation between exit availability and the potential for harm reduction during ASIs. This result aligns with previous findings that a greater number of exits enhances evacuation outcomes particularly on lower floors (Kodur et al., 2020).

For the impact of each exit, significant differences were seen among the subcases in terms of both evacuation rate ($F(6, 693) = 13.233$, $p < 0.001$, $\eta_p^2 = 0.103$, $d = 0.678$) and harm rate (F (6, 693) = 2.308, p = 0.033, $\eta_p^2$ = 0.020, d = 0.286). As shown in Figure 6, which is sorted in descending order by evacuation rate, blocking exits 1, 2, or 5 results in a higher increase in harm rate and a more significant reduction in evacuation rate compared to blocking exits 3, 4, or 6. Exits 1, 2 and 5 appear to be more critical related to their proximity to the shooter's initial location at the entrance. The closeness to the threat means these exits are vital for rapid evacuation. The result is consistent with fire safety research where the availability and location of exits are decisive for survival (Kobes et al., 2010). In the context of ASIs, the shooter's position relative to exit locations influences the effectiveness of evacuation routes and emergency exits. Maintaining clear access to these crucial exits, particularly near areas where shooting is likely to occur, is crucial to reduce harm and facilitate evacuation.

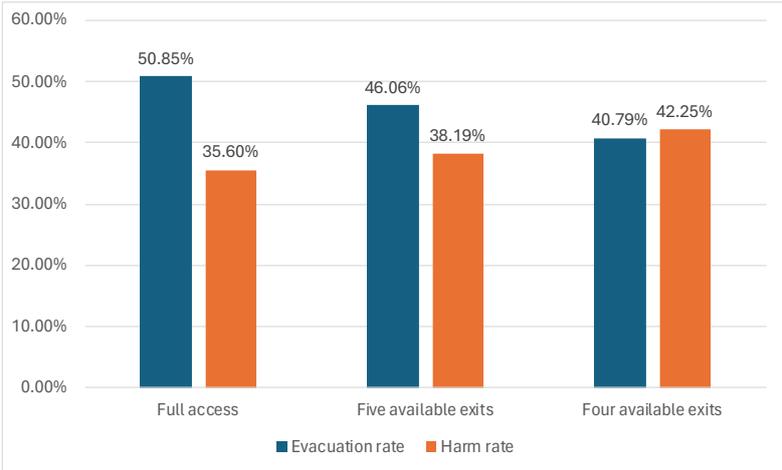

Figure 5: Significant Effects of Number of Exits

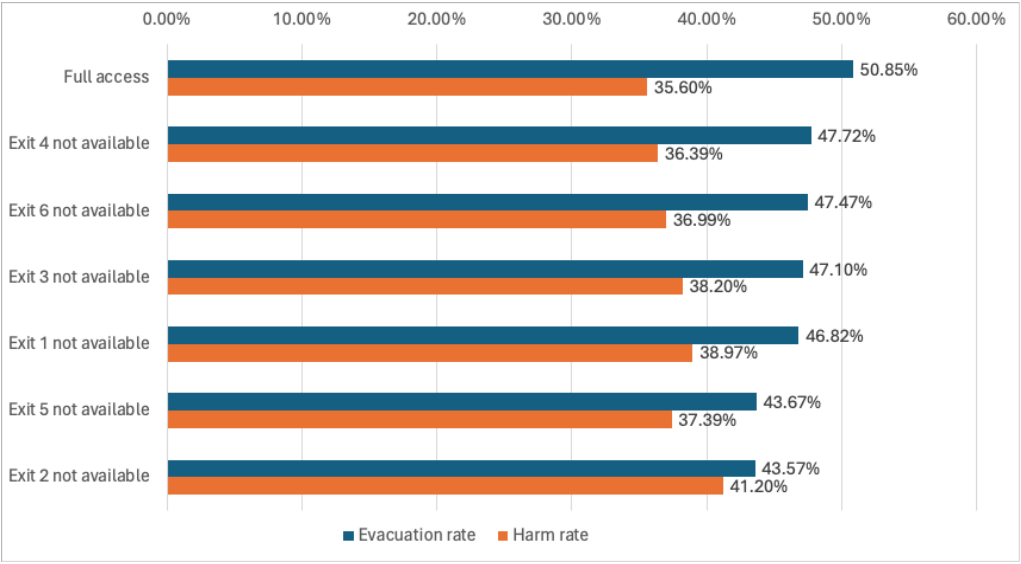

Figure 6: Evacuation Rate and Harm Rate of the Full Access and Five Available Exits Scenarios

Upon examination of scenarios with four available exits, we found significant differences among exit configurations on both evacuation rate (F (14, 1485) = 31.694, p < 0.001, $\eta_p^2$ = 0.230, d = 1.093) and harm rate (F (14, 1485) = 3.622, p < 0.001, $\eta_p^2$ = 0.033, d = 0.370). Table 2 arranges them in descending order based on the evacuation rate. The results show that blocking both Exit 5 and Exit 6 simultaneously is the most harmful configuration, leading to the highest harm rate and the lowest evacuation rate. Blocking exits 5 and 6 could disrupt the evacuation routes for a significant portion of the occupants as these exits serve the larger bottom wing of the building. Also, with these exits blocked, more building occupants may crowd into hiding places, causing overcrowding in those areas. The findings on exit configuration supports previous research indicating that the spatial arrangement of exits significantly affects evacuation outcomes (Kurdi et al., 2018; Lei and Tai, 2019). Yet, these prior investigations often employed simplified building layouts with fewer exits. This study extends beyond establishing generalized principles for exit availability, providing case-specific analyses to evaluate safety in more complex and realistic built environments.

Table 2: Evacuation Rate and Harm Rate of the Four Available Exits Scenarios

| Exit configuration | Evacuation rate (%) | Harm rate (%) |
|---|---|---|
| Exit 4 and Exit 6 not available | 44.99 $\pm$ 6.81 | 39.77 $\pm$ 14.02 |
| Exit 3 and Exit 4 not available | 44.65 $\pm$ 5.39 | 40.74 $\pm$ 13.04 |
| Exit 3 and Exit 6 not available | 44.53 $\pm$ 6.45 | 41.72 $\pm$ 12.39 |
| Exit 1 and Exit 4 not available | 44.41 $\pm$ 6.74 | 39.06 $\pm$ 11.12 |
| Exit 1 and Exit 6 not available | 43.07 $\pm$ 7.96 | 40.90 $\pm$ 11.23 |
| Exit 1 and Exit 3 not available | 42.65 $\pm$ 8.37 | 39.81 $\pm$ 10.54 |
| Exit 1 and Exit 5 not available | 41.21 $\pm$ 6.38 | 43.87 $\pm$ 12.16 |
| Exit 3 and Exit 5 not available | 40.84 $\pm$ 5.64 | 41.06 $\pm$ 14.97 |
| Exit 2 and Exit 4 not available | 40.66 $\pm$ 5.25 | 41.71 $\pm$ 14.73 |
| Exit 2 and Exit 3 not available | 40.22 $\pm$ 5.19 | 44.62 $\pm$ 12.69 |
| Exit 2 and Exit 6 not available | 40.15 $\pm$ 8.66 | 41.08 $\pm$ 15.25 |
| Exit 4 and Exit 5 not available | 39.32 $\pm$ 6.61 | 40.21 $\pm$ 14.41 |
| Exit 2 and Exit 5 not available | 37.49 $\pm$ 5.70 | 45.71 $\pm$ 15.39 |
| Exit 1 and Exit 2 not available | 36.05 $\pm$ 5.59 | 46.21 $\pm$ 11.81 |
| Exit 5 and Exit 6 not available | 31.66 $\pm$ 5.54 | 47.26 $\pm$ 18.97 |

A limitation of this study is the simplified assumptions regarding occupants' behavior including fixed pre-evacuation time and constant speed. Future studies could incorporate more dynamic decision-making that considers personal factors like training and exit familiarity, causing more diverse evacuation behaviors in real-life scenarios.

## 5. Conclusion

This study quantitatively assessed building safety design for ASIs, focusing on the number and configuration of exits, through RL-based simulations that incorporate dynamic shooter behaviors. Our case study findings demonstrate that the number and the configuration of exits significantly influence occupant safety, as evidenced by variations in harm and evacuation rates. Specifically, we find that a greater number of exits correlates with enhanced evacuation effectiveness and reduced occupant harm, emphasizing the importance of architectural design in enhancing public safety during ASIs. Also, the analysis of exit configurations illustrates that certain exit, especially those closer to the shooter's initial location, are more important for reducing harm and facilitating evacuation. This finding emphasizes the significance of strategic exit placement in building design.

This study extends previous studies by providing a case-specific examination of how various exit configurations can affect safety outcomes. The RL-based simulation method introduced in this study offers a tool for evaluating the effectiveness of different design strategies in mitigating ASI impacts. Future research could incorporate different environmental factors, shooter objectives, and occupants' behaviors to further validate the robustness of this simulation-based approach.

**Acknowledgement**

This material is based upon work supported by the U.S. Department of Homeland Security under Grant Award 22STESE00001-03-02. Dr. Lucas' effort is also funded by the Army Research Office and was accomplished under Cooperative Agreement Number W911NF-20-2-0053. The views and conclusions contained in this document are those of the authors and should not be interpreted as representing the official policies, either expressed or implied, of the Army Research Office, U.S. Department of Homeland Security, or the U.S. Government. The U.S. Government is authorized to reproduce and distribute reprints for Government purposes notwithstanding any copyright notation herein.